\documentclass[10pt,twocolumn,letterpaper]{article}

\usepackage{amsmath,graphicx}
\usepackage{amsmath}
\usepackage{amssymb}
\usepackage{booktabs}
\usepackage{multirow}
\usepackage[ruled,lined,linesnumbered]{algorithm2e}
\usepackage{balance}
\usepackage{subfigure}
\usepackage{balance}
\usepackage{url}
\usepackage[table]{xcolor}

\usepackage{wacv}
\usepackage{times}
\usepackage{epsfig}
\usepackage{graphicx}
\usepackage{amsmath}
\usepackage{amssymb}

\wacvfinalcopy % *** Uncomment this line for the final submission

\def\wacvPaperID{249} % *** Enter the wacv Paper ID here

\ifwacvfinal\fi
\begin{document}

%%%%%%%%% TITLE
\title{Pruning Deep Networks using Partial Least Squares}

%\title{Pruning Deep Neural Networks based on Feature Selection}% \\ 
%Partial Least Squares and Variable Importance in Projection}

% Authors at the same institution
%\author{First Author \hspace{2cm} Second Author \\
% Authors at different institutions
\author{Artur Jordao, Ricardo Kloss, Fernando Yamada, William Robson Schwartz \\
Smart Surveillance Interest Group, Computer Science Department\\
Universidade Federal de Minas Gerais, Brazil\\
Email: \{arturjordao, rbk,  fernandoakio, william\}@dcc.ufmg.br
}

\maketitle
\ifwacvfinal\thispagestyle{empty}\fi

\begin{abstract}
Modern pattern recognition methods are based on convolutional networks since they are able to learn complex patterns that benefit the classification. However, convolutional networks are computationally expensive and require a considerable amount of memory, which limits their deployment on low-power and resource-constrained systems.
To handle these problems, recent approaches have proposed pruning strategies that find and remove unimportant neurons (i.e., filters) in these networks. Despite achieving remarkable results, existing pruning approaches are ineffective since the accuracy of the original network is degraded.
%In this work, we propose a novel approach to efficiently remove filters in convolutional networks based on Partial Least Squares (PLS) and Variable Importance in Projection (VIP). These techniques allow estimating the filter importance based on its relationship with the class label, on a low-dimensional space, which we show to be an adequate indicator to remove filters. 
In this work, we propose a novel approach to efficiently remove filters from convolutional networks. Our approach estimates the filter importance based on its relationship with the class label on a low-dimensional space. This relationship is computed using Partial Least Squares (PLS) and Variable Importance in Projection (VIP).
Our method is able to reduce up to $67\%$ of the floating point operations (FLOPs) without penalizing the network accuracy. With a negligible drop in accuracy, we can reduce up to $90\%$ of FLOPs.
Additionally, sometimes the method is even able to improve the accuracy compared to original, unpruned, network. We show that employing PLS+VIP as the criterion for detecting the filters to be removed is better than recent feature selection techniques, which have been employed by state-of-the-art pruning methods. Finally, we show that the proposed method achieves the highest FLOPs reduction and the smallest drop in accuracy when compared to state-of-the-art pruning approaches. Codes are available at: https://github.com/arturjordao/PruningNeuralNetworks
\end{abstract}
\section{Introduction}\label{sec:introduction}
Convolutional networks have been an active research topic in Computer Vision mostly because they have achieved state-of-the-art results in numerous tasks~\cite{He:2016, Liu:2015}. However, convolutional networks are computationally expensive, present a large number of parameters and consume a considerable amount of memory, hindering applicability on low-power and real-time systems. To handle these problems, there exist three groups of approaches: (i) $1\times1$ convolutional filters, which reduce the dimensionality of the input feature map by squeezing the depth variables, decreasing the number of parameters~\cite{He:2016}; (ii) binarization of weights and activations, which replaces arithmetic operations with bitwise operations, improving speed-up and memory requirements~\cite{Hubara:2016, Rastegari:2016}; and (iii) pruning approaches, which remove neurons from a network, providing all the benefits of (i) and (ii) to deep architectures~\cite{Hu:2016, Li:2017, Huang:2018}. Based on these advantages, most efforts have focused on pruning methods.

Despite being simple and presenting considerable results, modern pruning approaches either require human effort or demand a high computational cost. In addition, current pruning criteria are ineffective since the accuracy of the original, unpruned, network is degraded. For instance, the method proposed by Li et al.~\cite{Li:2017} employs the L1-norm to locate candidate neurons (i.e., filters) to be eliminated. However, it requires considerable human effort to evaluate different tradeoffs between network performance and pruning rate (percentage of filters removed). Based on this limitation, Huang et al.~\cite{Huang:2018} proposed a pruning approach that removes unnecessary filters by learning pruning agents. These agents take the filter weights from a layer as input and output binary decisions indicating whether a filter will be kept or removed. Even though Huang et al.~\cite{Huang:2018} achieved superior performance to the hand-crafted pruning criterion by Li et al~\cite{Li:2017}, their method demands a higher computational cost because each agent is modeled as a neural network. In addition, when a higher number of the filters are eliminated the network accuracy decreases considerably.
%Also, their method requires several fine-tuning stages since the pruning is executed layer-by-layer.
\begin{figure*}[!htb]
	\centering
	\subfigure[] {\includegraphics[scale=0.5]{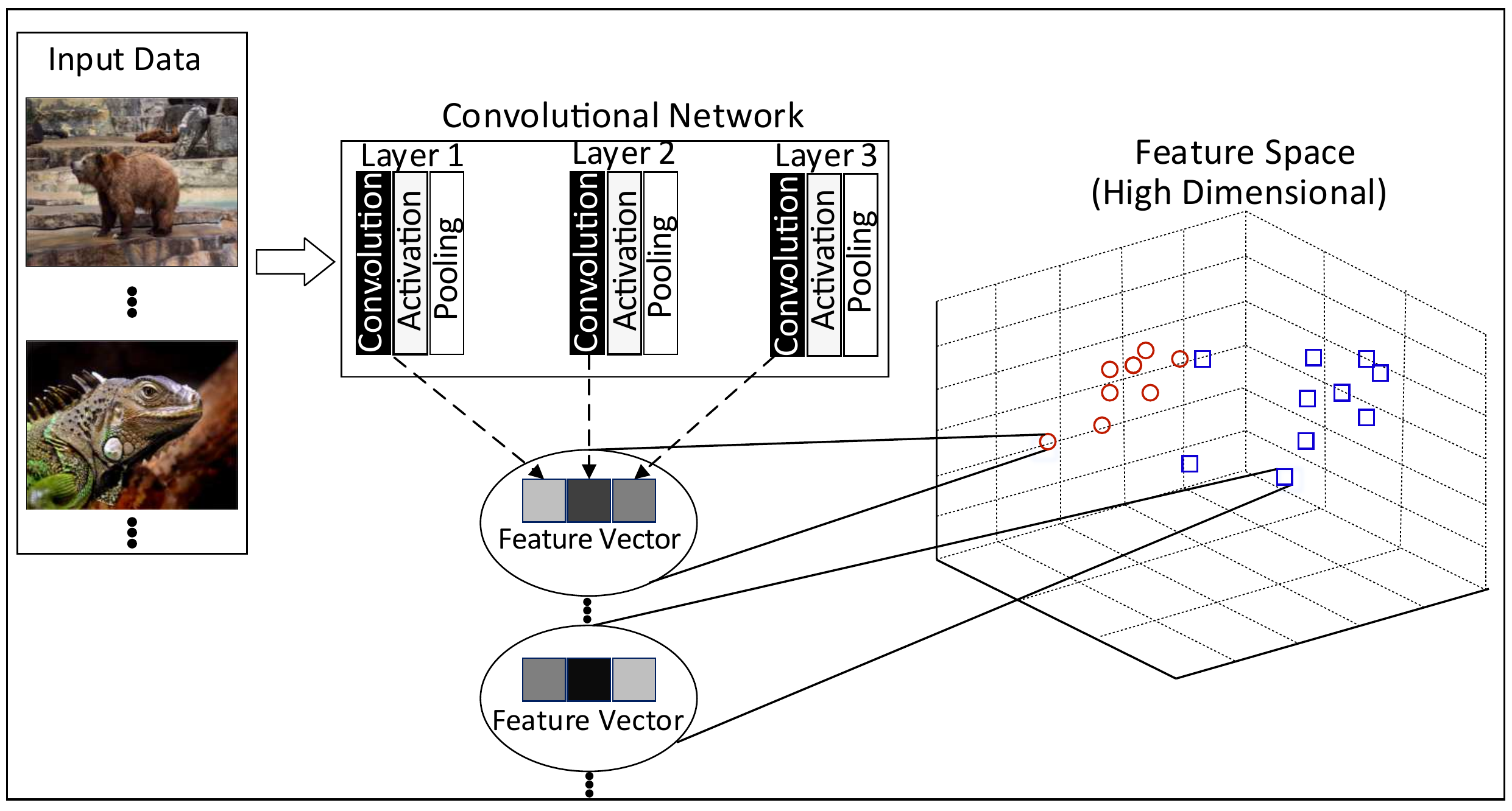}} 
	\hspace{10pt}
	\subfigure[] {\includegraphics[scale=0.54]{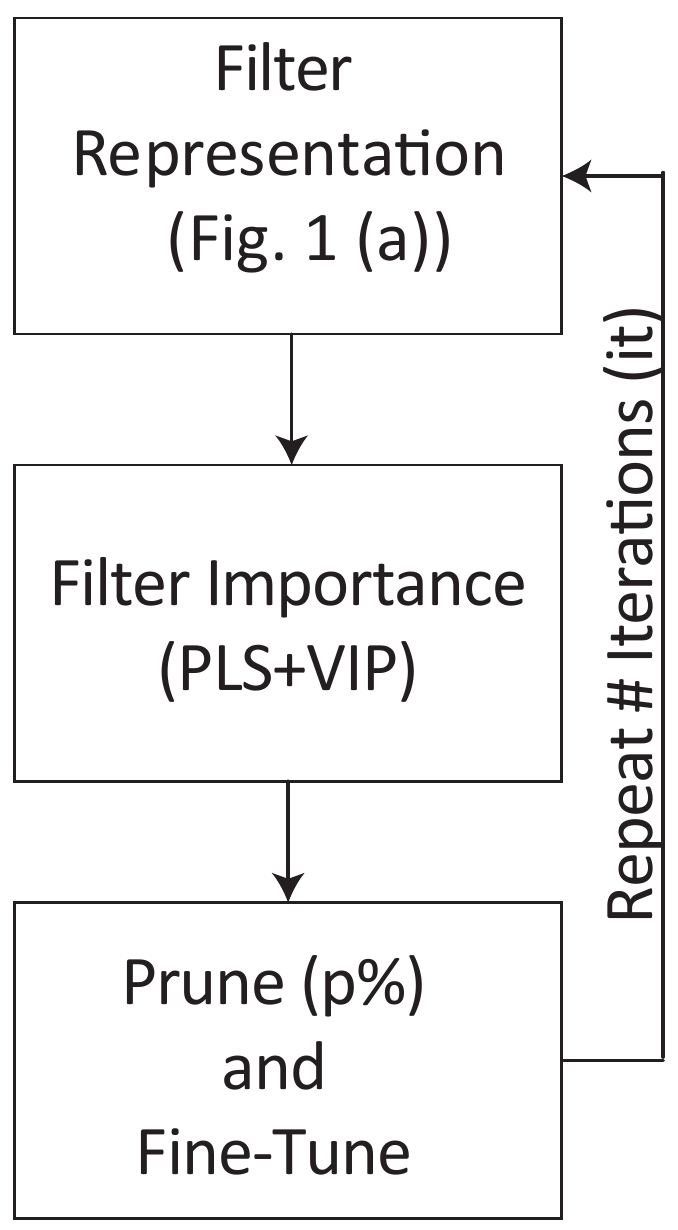}}	
	\caption{(a) Representation of convolutional filters as feature vectors. For simplicity, the network shows only one filter (one dimension of the feature space) in each layer. (b) Overview of the proposed method to prune convolutional filters from deep networks.}
	\label{fig:filter_representation}
\end{figure*}

Motivated by the limitations in current pruning methods, we propose a novel approach to efficiently eliminate filters in convolutional networks. Our method relies on the hypothesis that estimating the filter importance based on its relationship with the class label, on a low-dimensional space, is adequate to locate unimportant filters. This relationship is captured using Partial Least Squares, a discriminative feature projection method~\cite{Wold:1985, Mehmood:2012}.
An overview of our method is the following. First of all, we represent the convolutional filters of the network as features. To this end, we present the training data to the network and interpret the output of each convolutional filter as a feature vector (or a set of features), as illustrated in Figure~\ref{fig:filter_representation} (a). After this stage, we create a high dimensional feature space, representing all convolutional filters of the network at once. Then, we project this high dimensional space onto a latent space using Partial Least Squares (PLS). Next, we employ Variable Importance in Projection (VIP) to estimate the contribution of each feature in generating the latent space, enabling PLS to operate as a feature selection method. The idea behind this process is that, since the filters (i.e., its outputs) are represented as features, we are estimating the filter importance with respect to its relationship with the class label on the latent space (PLS criterion). Finally, we eliminate filters with low importance. This process can be iteratively repeated until a specific number of iterations, as illustrates Figure~\ref{fig:filter_representation} (b).

Different from existing pruning criteria, by using filter importance based on the PLS projection is extremely effective, where the method achieves the lowest drop in accuracy (sometimes improving the accuracy). In addition, PLS+VIP presents superior results than state-of-the-art feature selection techniques, which have been used to prune networks.
%
%	
%Furthermore, our pruning strategy examines all filters of the network at once, which enables us to eliminate filters coming from different layers simultaneously. 

We evaluate our method by pruning VGG16~\cite{Liu:2015} and ResNet~\cite{He:2016} on ImageNet~\cite{imagenet} and CIFAR-10~\cite{cifar} datasets, where we are able to reduce up to $67\%$ of FLOPs without penalizing network accuracy. With a negligible drop in accuracy, we can reduce up to $90\%$ of FLOPs. Furthermore, sometimes, the method improves network accuracy. 
%Experimental results show that using filter importance based on the PLS projection as the criterion for detecting the filters to be removed is extremely effective and better than recent feature selection techniques~\cite{Roffo:2015, Roffo:2017}, which have been used to prune neural networks.
%\input{Sections/RelatedWork.tex}
\section{Proposed Approach}\label{sec:proposed_method}
%This section defines our method to eliminate filters in convolutional networks. We start by describing the representation of \todo{convolutional filters} as feature vectors. Then, we explain Partial Least Squares and Variable Importance in Projection, which project the filter representation onto a low dimensional space and measure the filter importance, respectively. Finally, we describe how to remove filters with low importance. An overview of our method is presented in Figure~\ref{fig:pipeline}.

\noindent\textbf{Filter Representation.}
The first step in our method is to represent the output of the filters (i.e., its feature maps) that compose the network as feature vectors. For this purpose, let us consider we have $m$ training samples, which are forwarded on the network to obtain the feature maps provided by each convolutional filter. Since these feature maps are high dimensional, we apply a pooling operation to reduce their dimension. In this work, we consider the following pooling operations: global max and average pooling, and max-pooling $2\times2$. Finally, the output of the pooling operation is interpreted directly as one feature (when using the global pooling operations) or a set of features (when using the max-pooling $2\times2$). Specifically, each filter is represented by its feature maps followed by the pooling operation, as illustrates Figure~\ref{fig:filter_representation} (a).
{The intuition for using the feature map as a feature is that we are able to measure its relationship with the class label on the latent space (via PLS). In this way, a filter associated with a feature with low relationship might be removed.}

\noindent\textbf{Feature Projection.} After executing the previous step, we have generated a high dimensional space $R^d$ that represents all filters of the convolutional network. The second step of our method is to project this high dimensional space onto a low dimensional space $R^c$ ($c \ll d$), referred to as latent space. To this end, we employ Partial Least Squares (PLS), a discriminative feature projection method widely employed to model the relationship between dependent and independent variables. PLS works as follows. Let $X \subset R^{m \times d}$ and $y \subset R^{m \times k}$ be a matrix of independent and dependent variables, respectively. In our method, the matrix $X$ is the representation of the filters we have generated (first step of the proposed method) and $y$ is the class label matrix, where $k$ denotes the number of categories.

PLS estimates a projection matrix $W$ ($w_1, w_2, ...w_c$) that projects the high dimensional space $R^d$ onto a low dimensional space $R^c$ ($c$ is a parameter) such that each component $w_{i} \in W$ represents the maximum covariance between the $X$ and $y$, as shown in Equation~\ref{eq:pls}.
\begin{equation}\label{eq:pls}
w_i = argmax(cov(Xw, y)), \mathtt{s.t} \lVert w \rVert = 1.
\end{equation}
To solve Equation~\ref{eq:pls}, we can use Nonlinear Iterative Partial Least Squares (NIPALS) or SVD. In this work, we use NIPALS since it is faster than SVD. In addition, it allows us to find only the first $c$ components, while SVD finds all $d$ components, spending more computational resources. Algorithm~\ref{alg:nipals} introduces the steps of NIPALS to obtain the first $c$ components, where the convergence step is achieved when no changes occur in $w_i$. Also, we might define a number of steps as convergence criterion, to ensure the method stops.
\begin{algorithm}[!t]
	\small
	\caption{NIPALS Algorithm.}
	\label{alg:nipals}
	\SetKwInOut{Input}{Input}
	\SetKwInOut{Output}{Ouput}
	\Input{$X \subset \mathbb{R}^{m \times d}$, $y \subset \mathbb{R}^{m \times k}$}
	\Input{Number of components $c$}
    \Output{$W \subset \mathbb{R}^{d \times c}$}
	\BlankLine
	\For{ $i=1$ \textbf{to} $c$}{
		randomly initialize $u \in \mathbb{R}^{m \times 1}$
		\BlankLine
		$w_i = \frac{X^Tu}{\lVert X^Tu \rVert}$, where $w_i \in W$
		\BlankLine
		$t_i = Xw_i$, 	$q_i = \frac{y^Tt_i}{\lVert y^Tt_i\rVert}$
		\BlankLine
		$u = yq_i$
		\BlankLine
		Repeat steps $3-5$ until convergence
		\BlankLine
		%$p_i = 
		\BlankLine
		$X = X - t_i(X^Tt_i)^T$, $y = y - t_iq_i^T$
	}
\end{algorithm}

Note that, in this step of our method, other feature projection methods could be employed, e.g., Principal Component Analysis (PCA) or Linear Discriminant Analysis (LDA). However, we believe that the idea behind PLS, which is to capture the relationship between the feature (in our context a filter) and its class label, is more suitable. In particular, when compared to LDA, PLS is robust to sample size problem (singularly)~\cite{Martinez::2001}. Moreover, PLS can be learned using few samples, not requiring all the data to be available in advance. These advantages make PLS more flexible and efficient than traditional feature projection methods, mainly for large datasets and resource-constrained systems.

\noindent\textbf{Filter Importance.} The next step in our method is to measure the filter importance score to remove the ones with low importance. To this end, once we have found the projection matrix $W$, we estimate the importance of each feature based on its contribution to yield the latent space employing the Variable Importance in Projection (VIP) technique~\cite{Mehmood:2012}. For each feature, $f_j$, VIP calculates its importance in terms of
\begin{equation}\label{eq:vip}
f_j = \sqrt{d\sum_{i=1}^c S_i(w_{ij}/\Vert w_i \Vert ^2)/\sum_{i=1}^c S_i},
\end{equation}
where $S_i$ is the sum of squares explained by the i-\emph{th} component, which can alternatively be expressed as $q_i^2t'_it_i$ (defined in Algorithm~\ref{alg:nipals})~\cite{Mehmood:2012}. We highlight that following the modeling performed in the first step of our method, a feature is associated (i.e. correspond) with a filter. This is because the feature is represented by the feature maps of the filter.
Observe that when using the max-pooling operation as filter representation, we have a set of features for each filter; in this case, the final score of a filter is the average of its $f_j$.

\noindent\textbf{Prune and Fine-tuning.}
Given the importance of all filters that compose the network, we have generated a set of scores, $\{f_1, f_2,..., f_j\}$. Then, given a pruning ratio $p$ (e.g., $10\%$), we remove $p\%$ of the filters {based on its scores}.
%The value of the threshold can be found by analyzing the percentage of filters removed if we consider $f_k$ as $th$ and select the one which better approximates the desired $p$.
By executing all these steps, we have executed one iteration of the proposed method, as illustrated in Figure~\ref{fig:filter_representation} (b). Note that the input network to the next iteration is the pruned network of the previous iteration.
\section{Experimental Results}\label{sec:experiments}
\noindent\textbf{Experimental Setup.} We conduct experiments using a single NVIDIA GeForce GTX 1080 TI on a machine with 64GB of RAM. Following previous works~\cite{Li:2017, Huang:2018}, we examine some aspects and parameters of our method by considering VGG16 only on CIFAR-10 and discuss the results using drop in accuracy in percentage points (p.p), where negative values denote improvement w.r.t the original, unpruned, network. Also, we compute FLOPs following the work of Li et al.~\cite{Li:2017}. Finally, we set the pruning rate to $10\%$ and the number of components of PLS to $c=2$, where our method resulted in the least drop in accuracy.
\begin{figure}[!b]
	\centering
	\includegraphics[scale=0.5]{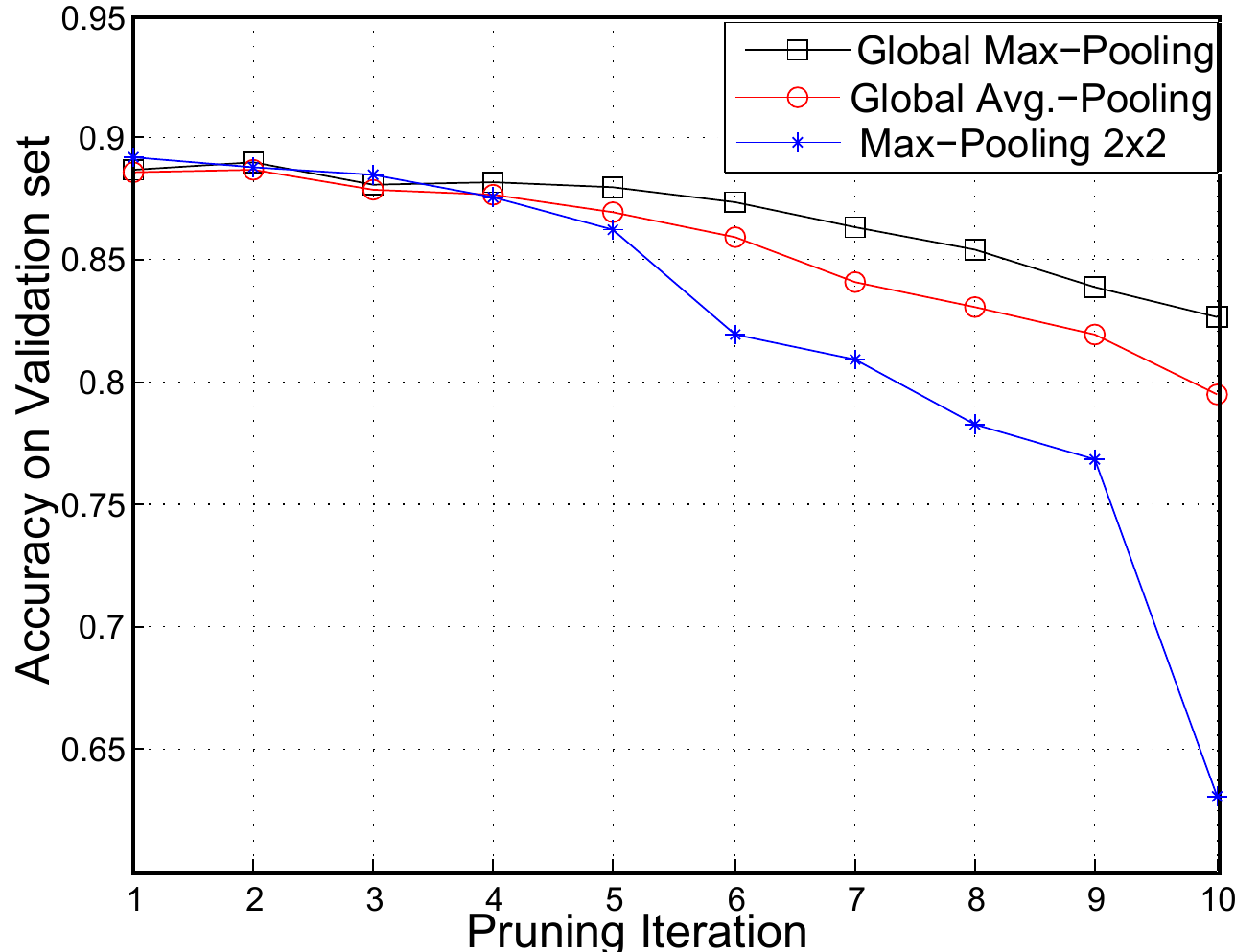}
	\caption{Accuracy obtained by pruning VGG16 using different filter representations.}
	\label{fig:comparisonrepresentationvalidation}
\end{figure}

\noindent\textbf{Influence of the filter representation.} One of the most important issues in our method is the pooling operation, referred to as filter representation, employed on the feature map provided by a filter. This experiment aims at validating this issue. For this purpose, we execute ten pruning iterations using different pooling operations. As illustrated in Figure~\ref{fig:comparisonrepresentationvalidation}, accuracy decreases slower when global max-pooling is employed. On the contrary, by using the max-pooling $2\times2$ accuracy drops faster, where at the $10$th iteration the method drops $26$ p.p. compared to the unpruned network. In addition, this representation has the drawback of consuming additional memory compared to the global operations which reduce the feature map to one dimension. Based on this result, we use the global max-pooling as filter representation in the remaining experiments.

\noindent\textbf{Number of samples to learn the PLS.} 
%A basic requirement in deep learning approaches is a large number of training samples to avoid overfitting and provide a good generalization. Based on this statement, large datasets have been proposed, e.g., ImageNet~\cite{imagenet} with $1.2$ million images. 
On large datasets, our method could be impracticable due to memory constraints, since NIPALS requires all the samples be in memory. However, an advantage of PLS is that it can be learned with a small number of samples. Thus, we can subsample $X$ before executing NIPALS, enabling our method to operate on large datasets.

In this experiment, we intend to demonstrate that the proposed method is robust when fewer samples are used to learn the PLS. To this end, we vary the percentage of training samples (using a uniform subsampling) used to compose $X$ in Algorithm~\ref{alg:nipals}. Table~\ref{tab:number_samples} shows the results obtained after one pruning iteration, where it is possible to observe that the network accuracy is slightly changed as a function of the number of samples used to learn the PLS. In particular, sometimes, the accuracy is the same as employing $100\%$ of the samples (e.g., using $20\%$). Also, the difference between using $100\%$ and $10\%$ of the samples is only $0.1$ p.p.. Thus, to conduct the experiments on ImageNet, we used only $10\%$ of the samples.
\begin{table}[!t]
	\centering
	\caption{Accuracy by pruning VGG16 on CIFAR-10 (validation set), using different number of samples to learn the PLS.}
	\label{tab:number_samples}
	\begin{tabular}{cc}
		\hline
		($\%$)Training Samples & Accuracy after Pruning\\ \hline
		$10 $& $89.7$ \\
		$20 $& $89.8$ \\
		$40 $& $89.7$ \\
		$60 $& $89.8$ \\
		$80 $& $89.6$ \\
		$100$ & $89.8$ \\ \hline
	\end{tabular}
\end{table}

\noindent\textbf{Iterative pruning vs. single pruning.} In this experiment, we show that it is more appropriate to execute our method iteratively, as illustrated in Figure~\ref{fig:filter_representation} (b), with a low pruning ratio (i.e $10\%$) instead of using a single pruning iteration with a high pruning ratio. In other words, if we want to remove i.e. $40\%$ of filters, it is better to execute some iterations of our method with a low pruning ratio instead of setting a pruning ratio of $40\%$ and execute only a single iteration. For this purpose, we first execute five iterations of the proposed method with a pruning ratio of $10\%$. Then, after each iteration, we compute the percentage of removed filters, $p_i$. Finally, we use each $p_i$ as the pruning ratio to execute a single iteration of the method. According to the results in Table~\ref{tab:iterative_single}, performing our method iteratively with a low pruning ratio is more effective than using it with a large pruning ratio, which led to a higher drop in accuracy. For instance, by executing five iterations of the method with a pruning ratio of $10\%$, we are able to remove $40\%$ of filters while improving the network accuracy (indicated by negative values in Table~\ref{tab:iterative_single}). On the other hand, by applying a single iteration with a pruning ratio of $40\%$, the accuracy decreased $1.76$ p.p..
\begin{table}[!t]
	\centering
	\caption{Drop in accuracy when executing our method with few iterations and a low pruning ratio (Iterative Pruning), and when executing a single iteration with a high pruning ratio (Single Pruning). Results on CIFAR-10 (test set).}
	\label{tab:iterative_single}
	\begin{tabular}{ccc}
		\hline
		\begin{tabular}[c]{@{}c@{}}Percentage of \\ Removed Filters\end{tabular} & \begin{tabular}[c]{@{}c@{}}Iterative Pruning \\ Accuracy$\downarrow$\end{tabular} & \begin{tabular}[c]{@{}c@{}}Single Pruning \\ Accuracy$\downarrow$\end{tabular} \\ \hline
		$10$ & $-0.89$ (it=1) & $-0.89$ \\
		$27$ & $-1.08$ (it=3) & $-0.03$ \\
		$40$ & $-0.69$ (it=5) & $1.76$ \\
		$65$ & $1.56$ (it=10) & $20.21$ \\ \hline
	\end{tabular}
\end{table}

\noindent\textbf{Comparison with other pruning criteria.}
{The idea behind this experiment is to demonstrate that the criterion employed by our method is more effective to eliminate filters than existing pruning criteria as well as state-of-the-art feature selection techniques.} To this end, we use one iteration of pruning and follow the process suggested in~\cite{Yu:2018}, which consists of setting the same pruning ratio ($10\%$) and modifying only the criterion to select the filters to be removed. By employing one pruning iteration, we are able to show the robustness of the methods when employing a single stage of fine-tuning. Recall that, for each iteration of our method, we execute a single stage of fine-tuning, thereby, the number of iterations defines the number of fine-tuning stages. Finally, as input to the methods of feature selection (Infinity FS~\cite{Roffo:2015} and Infinity Latent FS~\cite{Roffo:2017}), we use the global max-pooling filter representation.
\begin{table}[!b]
	\centering
	\small
	\caption{Drop in accuracy using different criteria to determine the filter importance.}
	\label{tab:pruning_criterions}
	\begin{tabular}{cc}
		\hline
		Filter Importance Criterion & Accuracy drop$\downarrow$\\ \hline
		L1Norm~\cite{Li:2017} & $-0.69$\\
		APoZ~\cite{Hu:2016} & $-0.70$\\ 
		Infinity FS~\cite{Roffo:2015} & $-0.69$\\
		Inifinity Latent FS~\cite{Roffo:2017} & $-0.65$\\ \cline{1-2} 
		PLS+VIP (Ours) & $\mathbf{-0.89}$ \\
		\hline
	\end{tabular}
\end{table}

Table~\ref{tab:pruning_criterions} shows the results obtained by different pruning criteria. According to the results, our criterion to define the filter importance is more suitable than L1-norm and APoZ, where we achieve the lower drop in accuracy. In addition, PLS+VIP achieved superior performance when compared to methods designed specifically for feature selection~\cite{Roffo:2015, Roffo:2017}. 
The reason for these results is that PLS preserves filters with high relationship with the class label, which are the most important to the classification ability of the network.
\begin{table}[!t]
	\caption{ Comparison with state-of-the-art pruning approaches (results reported by the original papers). Acc.$\downarrow$ denotes drop in accuracy (in percentage points), where negative values denote improvement regarding the original, unpruned, network. FLOPs $\downarrow$ denotes the percentage of FLOPs reduced ({the higher the better}) w.r.t the original network.}
	\label{tab:results}
	\small
	\begin{tabular}{cccc}
		\hline
		& Method  & FLOPs$\downarrow$ & Acc.$\downarrow$ \\ \hline
		\multicolumn{1}{c|}{\multirow{6}{*}{\begin{tabular}[c]{@{}c@{}}VGG16 on \\ CIFAR-10\end{tabular}}}    & Hu et al.~\cite{Hu:2016}  & $28.29$ & $-0.66$ \\
		\multicolumn{1}{c|}{} & Li et al.~\cite{Li:2017} &  $34.00$ & $-0.10$ \\
		\multicolumn{1}{c|}{} & Huang et al.~\cite{Huang:2018}  & $64.70$ & $1.90$ \\ \cline{2-4} 
		\multicolumn{1}{c|}{} & Ours (it=1) & $23.13$ & $-0.89$ \\
		\multicolumn{1}{c|}{}                                                                                 & Ours (it=5)                       &  $67.25$           & $-0.63$              \\
		\multicolumn{1}{c|}{}                                                                                 & Ours (it=10)                       &  $90.66$           & $1.50$                \\ \hline
		%%%%%%%%%%%%%%%%%%%%%%%%%%%%%%%%%%%%%%%%%%%%%%%%%%%%%%%%%%%%%%%%%%%%%%%%%%%%%%%%%%%%%%%%%%%%%%%%%
		\multicolumn{1}{c|}{\multirow{7}{*}{\begin{tabular}[c]{@{}c@{}}ResNet56 on\\ CIFAR-10\end{tabular}}}  & Li (A)~\cite{Yu:2018}  & $10.40$        & $-0.06$              \\
		\multicolumn{1}{c|}{}  & Li (B)~\cite{Yu:2018}  & $27.60$           & $-0.02$              \\
		\multicolumn{1}{c|}{} & Yu et al.~\cite{Yu:2018} &  $43.61$ & $0.03$ \\ 
		\multicolumn{1}{c|}{} & He et al.~\cite{He:2018} &  $50.00$ & $0.90$ \\ \cline{2-4}
		\multicolumn{1}{c|}{} & Ours(it=1) &  $7.09$ & $-0.60$ \\
		\multicolumn{1}{c|}{} & Ours(it=5)  &  $35.23$  & $-0.90$\\
		\multicolumn{1}{c|}{} & Ours(it=8) &   $52.56$ & $-0.62$  \\ \hline
		%%%%%%%%%%%%%%%%%%%%%%%%%%%%%%%%%%%%%%%%%%%%%%%%%%%%%%%%%%%%%%%%%%%%%%%%%%%%%%%%%%%%%%%%%%%%%%%%%
		\multicolumn{1}{c|}{\multirow{7}{*}{\begin{tabular}[c]{@{}c@{}}ResNet110 on\\ CIFAR-10\end{tabular}}}  & Li (A)~\cite{Yu:2018}             & $15.90$          & $0.02$              \\
		\multicolumn{1}{c|}{} & Li (B)~\cite{Yu:2018}			  &  $38.60$           & $0.23$              \\
		\multicolumn{1}{c|}{}  & Yu et al.\cite{Yu:2018} &$43.78$ & $0.18$ \\ \cline{2-4}
		\multicolumn{1}{c|}{} & Ours(it=1) &  $6.85$ & $-0.59$ \\
		\multicolumn{1}{c|}{} & Ours(it=5)  &  $33.16$ & $-1.51$              \\
		\multicolumn{1}{c|}{} & Ours(it=7) &  $44.46$ & $-1.39$ \\ \hline
		%%%%%%%%%%%%%%%%%%%%%%%%%%%%%%%%%%%%%%%%%%%%%%%%%%%%%%%%%%%%%%%%%%%%%%%%%%%%%%%%%%%%%%%%%%%%%%%%%
		\multicolumn{1}{l|}{\multirow{8}{*}{\begin{tabular}[c]{@{}c@{}}VGG16 on\\ImageNet\\($224\times224$)\end{tabular}}}   & Li et al.~\cite{Li:2017} & $20.00$ & $14.60$              \\
		%#####Results of http://openaccess.thecvf.com/content_ICCV_2017/papers/Luo_ThiNet_A_Filter_ICCV_2017_paper.pdf
		%###########
		%\multicolumn{1}{l|}{} & He et al.~\cite{He:2017:ICCV} & $20.00$ & $1.7$ \\
		\multicolumn{1}{l|}{} & Wang et al.~\cite{Wang:2018} &  $20.00$ & $2.00$\\
		\multicolumn{1}{l|}{} & He et al.~\cite{He:2018} &  $20.00$ & $1.40$
		\\ \cline{2-4} 
		\multicolumn{1}{l|}{} & Ours(it=1) &  $9.31$ & $-0.98$\\
		\multicolumn{1}{l|}{} & Ours(it=3) &  $36.03$ & $1.06$\\
		\multicolumn{1}{l|}{} & Ours(it=5) &  $59.27$ & $2.21$ \\ \hline
	\end{tabular}
\end{table}

\noindent\textbf{Comparison with existing pruning approaches.} This experiment compares the proposed method with state-of-the-art pruning approaches. 
For this purpose, we report the results using one and five iterations of our method and the iteration where it achieved the closest drop in accuracy compared to the best method. Table~\ref{tab:results} summarizes the results.

On CIFAR-10, our method achieved the best tradeoff between the {drop/improvement} in accuracy and FLOPs reduction. When compared to Hu et al.~\cite{Hu:2016} and Li et al.~\cite{Li:2017}, we achieved around $2\times$ more FLOPs reduction with superior improvement in accuracy on both networks. Compared to Huang et al.~\cite{Huang:2018}, our method decreased $1.5\times$ more FLOPs with a smaller drop in accuracy on VGG16. 
In addition, by pruning ResNet56 and ResNet110, our method achieved a higher FLOPs reduction than the most recent pruning approaches~\cite{Yu:2018, He:2018}.
We also compared the proposed method with Li et al.~\cite{Li:2017} (A) and (B), which consists of employing several pruning ratios for different parts of the network. By performing five iterations of the proposed method, we outperformed Li (A) and (B) on both FLOPs reduction and accuracy improvement. Observe that, while existing pruning approaches degrade the network accuracy, our method is, in fact, able to  improve accuracy.

On ImageNet, with only three iterations of the proposed method, we were able to achieve the smallest drop in accuracy and $1.80\times$ more FLOPs reduction than all the methods. Also, with two additional iterations, we decreased about $3\times$ more FLOPs than all other methods.
\begin{figure}[!t]
	\centering
	\subfigure[] {\includegraphics[scale=0.314]{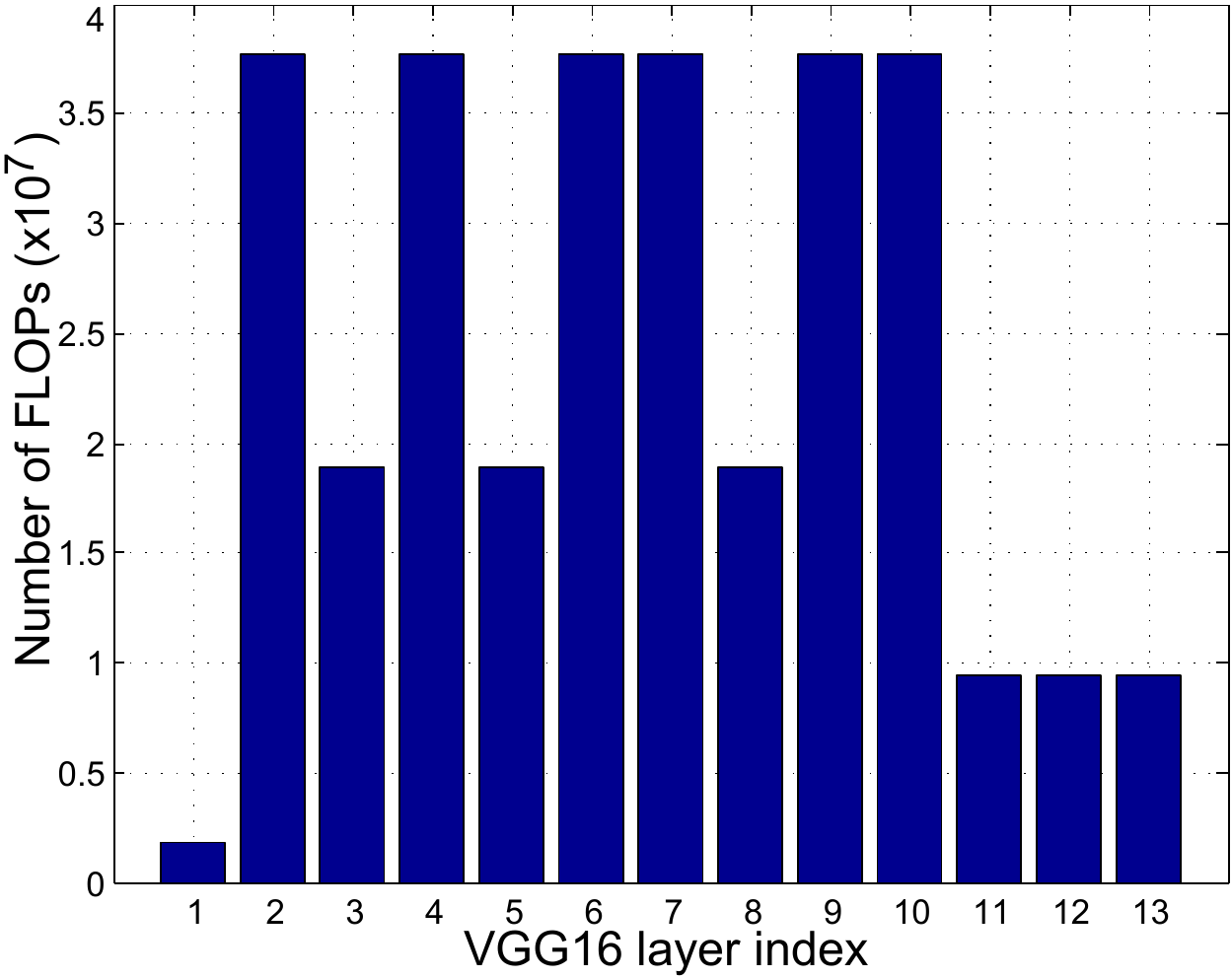}} 
	%	\hspace{-1pt}
	\subfigure[] {\includegraphics[scale=0.32]{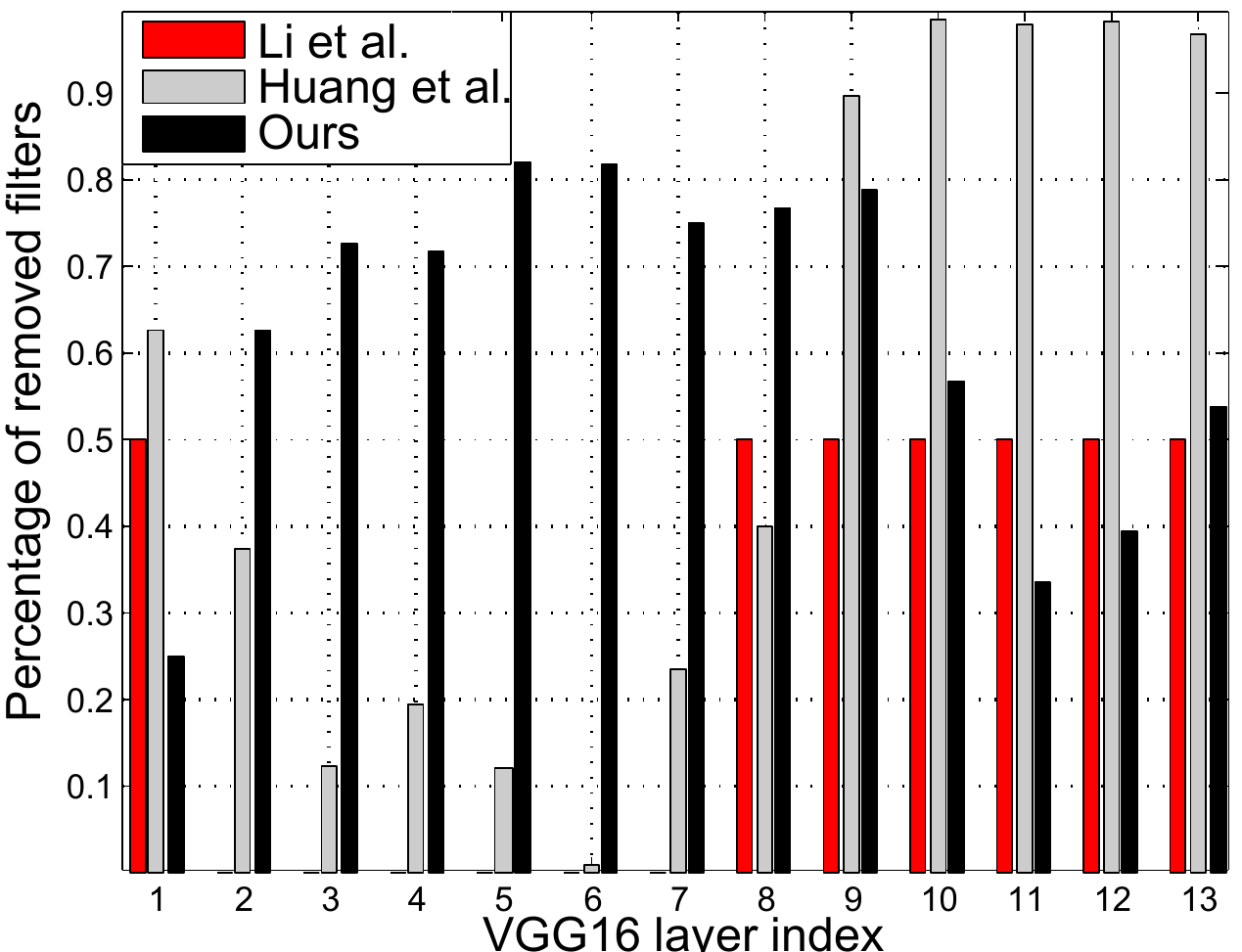}}
	\caption{(a) Number of float point operations (FLOPs) per layer of the VGG16 network. (b) Percentage of removed filters in each layer using different pruning methods.}
	\label{histogram_discard}
\end{figure}

Based on the aforementioned discussion, we have shown that the proposed method attains a superior reduction in FLOPs. This is an effect of the layers where it removes the filters. According to Figure~\ref{histogram_discard} (a), the layers $2$, $4$, $6$, $7$, $9$ and $10$ have the highest number of FLOPs. In general, the existing methods fail to eliminate filters from these layers. For instance, the methods proposed by Li et al.~\cite{Li:2017} and Huang et al.~\cite{Huang:2018} remove a large number of filters from the layers $9$ to $13$ (Figure~\ref{histogram_discard} (b)), but they remove a small number of filters from other layers. On the contrary, our method eliminates a large number of filters from all layers, as shown in Figure~\ref{histogram_discard} (b). In particular, we eliminate more than $50\%$ of filters from layers $2$ to $10$, which are the ones with the largest number of FLOPs, and more than $25\%$ from the other layers. Hence, we are able to achieve a higher FLOPs reduction than existing state-of-the-art methods, which are biased in eliminating filters of particular layers.
\section{Conclusions}\label{sec:conclusions}
This work presented an accurate pruning method to remove filters from convolutional networks.
The proposed method interprets each filter as a feature vector and creates a high dimensional space using these features. Then, it projects this space onto a low-dimensional latent space, using Partial Least Squares, {which captures the relation between the feature (filter) and its class label.} Finally, the method estimates the importance of each feature to yield the latent space and removes the ones with low importance {(low relationship with the class label).}
The method is able to reduce up to $67\%$ of FLOPs without penalizing the network accuracy. In particular, it is even able to improve the accuracy regarding the original, unpruned, network. In addition, with a negligible drop in accuracy, the method is able to reduce up to $90\%$ of FLOPs.
Compared to state-of-the-art pruning methods, the proposed method is extremely effective, where it attains the highest FLOPs reduction and the smallest drop in accuracy.

\section*{Acknowledgments}
The authors would like to thank the Brazilian National Research Council -- CNPq (Grants \#311053/2016-5 and \#438629/2018-3), the Minas Gerais Research Foundation -- FAPEMIG (Grants APQ-00567-14 and PPM-00540-17) and the Coordination for the Improvement of Higher Education Personnel -- CAPES (DeepEyes Project).

\balance
{\small
\bibliographystyle{ieee}
\bibliography{egbib}
}
\end{document}